\documentclass[conference]{IEEEtran}
\IEEEoverridecommandlockouts
\usepackage{cite}
\usepackage{amsmath,amssymb,amsfonts}
\usepackage{algorithmic}
\usepackage{graphicx}
\usepackage{textcomp}
\usepackage{xcolor}
\usepackage{subfig} 
\usepackage[hyphens]{url}
\usepackage{fancyhdr}
\usepackage{amsmath}
\usepackage{tablefootnote}
\def\BibTeX{{\rm B\kern-.05em{\sc i\kern-.025em b}\kern-.08em
    T\kern-.1667em\lower.7ex\hbox{E}\kern-.125emX}}
    
\usepackage{lipsum}
\newcommand\blfootnote[1]{%
  \begingroup
  \renewcommand\thefootnote{}\footnote{#1}%
  \addtocounter{footnote}{-1}%
  \endgroup
}

\begin{document}

\title{4bit-Quantization in Vector-Embedding for RAG}

\author{\IEEEauthorblockN{Taehee Jeong}
\IEEEauthorblockA{\textit{San Jose State University} \\
taehee.jeong@sjsu.edu}
}

\maketitle

\begin{abstract}
Retrieval-augmented generation (RAG) is a promising technique that has shown great potential in addressing some of the limitations of large language models (LLMs). 
LLMs have two major limitations: they can contain outdated information due to their training data, and they can generate factually inaccurate responses, a phenomenon known as hallucinations.
RAG aims to mitigate these issues by leveraging a database of relevant documents, which are stored as embedding vectors in a high-dimensional space. 
However, one of the challenges of using high-dimensional embeddings is that they require a significant amount of memory to store. This can be a major issue, especially when dealing with large databases of documents. To alleviate this problem, we propose the use of 4-bit quantization to store the embedding vectors. This involves reducing the precision of the vectors from 32-bit floating-point numbers to 4-bit integers, which can significantly reduce the memory requirements. Our approach has several benefits. Firstly, it significantly reduces the memory storage requirements of the high-dimensional vector database, making it more feasible to deploy RAG systems in resource-constrained environments. Secondly, it speeds up the searching process, as the reduced precision of the vectors allows for faster computation. Our code is available at https://github.com/taeheej/4bit-Quantization-in-Vector-Embedding-for-RAG.

\end{abstract}

\begin{IEEEkeywords}
Retrieval-augmented generation, Large language models, Vector search, Embedding vectors, Quantization, High-dimensional vectors, Memory storage, Vector database
\end{IEEEkeywords}


\section{Introduction}
\renewcommand{\footnoterule}{%
  \kern -3pt
  \hrule width 3in height 1pt
  \kern 2pt
}
\blfootnote{23rd International Conference on Machine Learning and Applications (ICMLA), 
IEEE Copyright 2024}

Large Language Models (LLMs) have achieved state-of-the-art performance on many Natural Language Processing (NLP) tasks \cite{b1, b2, b3}, demonstrating their ability to store a vast amount of knowledge as implicit parameters. The responses generated by LLMs, given a  query, are often useful and informative, as they are based on a wide variety of information learned during the training process. However, despite their impressive performance, LLMs still suffer from several limitations.

One major issue is that the number of model parameters required to achieve good performance is growing exponentially. This means that as the model is trained on more data, the number of model parameters required to store the learned knowledge increases rapidly, making it challenging to deploy and maintain these models. Another limitation is that LLMs are fundamentally limited in their ability to incorporate time-sensitive or not-publicly available information. This is because the data used to train LLMs is typically based on a snapshot of the internet at a particular point in time, and may not reflect the latest developments or updates.

As a result, LLMs are prone to producing fact hallucinations, where the model generates responses that are not grounded in reality. This can be particularly problematic in applications where accuracy and reliability are critical.

\begin{figure}[ht]
\centerline{\includegraphics[width=0.9\linewidth]{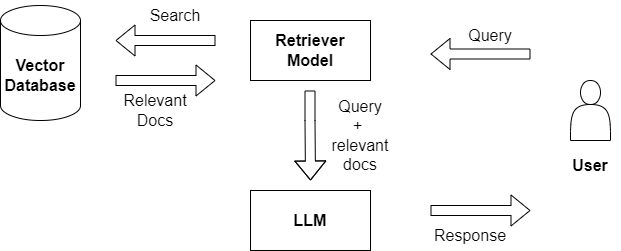}}
\caption{RAG system}
\label{fig:rag}
\end{figure}

Recently, a new paradigm has emerged to address these limitations: Retrieval-Augmented Generation (RAG). RAG combines pre-trained language models with external knowledge retrieval, allowing the model to access and incorporate time-sensitive and not-publicly available information \cite{b4, b5}. The RAG system works as follows: for a given input query, a neural retriever performs a search to find the most relevant documents from an external knowledge database. The retrieved documents, along with the query, are then used to prompt the language model, which generates its output by augmenting the retrieved documents with the given query. The work flow of the RAG system is illustrated in Fig. \ref{fig:rag}.

Compared to generative-only models, RAG explicitly exploits time-sensitive information and/or not-publicly available documents, such as a company's internal documents, and prevents potential hallucinations. Additionally, the language model in RAG requires a much smaller number of parameters, typically in the range of 7-8 billion, which is significantly smaller than the number of parameters required by generative-only models, such as Chat-GPT 3.5 \cite{b6}, which requires 175 billion parameters. The reduction in the number of parameters is available because the language model in RAG generates responses based on retrieved information from a vector database, rather than relying on implicit model parameters. This approach allows the model to access and incorporate a vast amount of knowledge, without requiring a large number of model parameters.

\begin{figure}[ht]
\centerline{\includegraphics[width=0.9\linewidth]{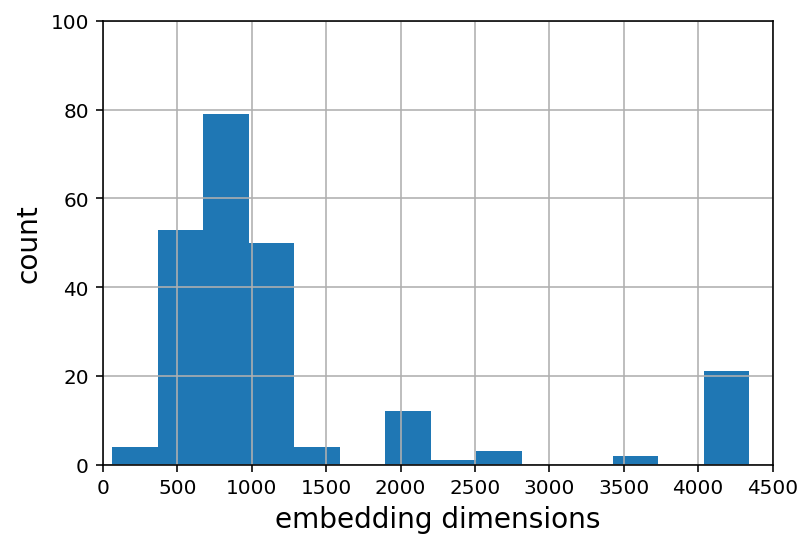}}
\caption{Distribution of embedding dimensions}
\label{fig:emb_dim}
\end{figure}

In RAG systems, relevant documents are represented as embedding vectors within a high-dimensional space. Notably, the top-ranked models on the Massive Text Embedding Benchmark (MTEB) leaderboard \cite{b7, b8}, as shown in Fig. \ref{fig:emb_dim}, consistently utilize embedding dimensions ranging from 512 to 4096.


However, high-dimensional embedding vectors require a large amount of memory to store. For example, the dbpedia-openai-1M-1536-angular dataset\cite{b9} requires $1M \times 1536 \times 4bytes =6.1GB$ of RAM to store the embedding vectors alone, not including the indexing and text data. To compress the embedding vectors, 8-bit quantization can be applied, reducing the memory requirements to $1M \times 1536 \times 1bytes =1.5GB$. Further compression to 4-bit integer can reduce the memory requirements to $1M \times 1536 \times 0.5bytes =0.75GB$.

In this work, we propose 8-bit or 4-bit quantization of embedding vectors, which can significantly reduce the memory storage and usage of a vector database. This approach can substantially reduce the cost of hosting a vector database on cloud services and is expected to speed up the searching process. However, the searching accuracy may be affected, and we will discuss this in detail.

\section{Related Work}
Traditional searching algorithms, such as bag-of-words retrieval methods, rank a set of documents based on the frequency of query terms appearing in each document. One popular algorithm is Term Frequency-Inverse Document Frequency (TF-IDF), which evaluates the importance of a word in a document relative to a collection of documents by multiplying the Term Frequency (TF) and Inverse Document Frequency (IDF) values. TF-IDF is widely used in search engines to rank documents based on their relevance to a user's query.

Another algorithm is Best Matching 25 (BM25) \cite{b10}, which is a variant of the TF-IDF algorithm. 
BM25 estimates a probability of relevance of documents to a given query by incorporating TF, IDF, and document length normalization.  BM25 uses a normalization factor to ensure that the scores are comparable across documents of different lengths, which makes effective at handling long documents. BM25 is a powerful and widely-used ranking function in information retrieval, known for its balance  between simplicity and effectiveness. 

Unlike traditional searching algorithms that look for exact matches of information like keyword matches, vector search represents data points as vectors, which have direction and magnitude, in a highly-dimensional space. With vector search, the individual dimensions define a specific attribute or feature, and the search compares the similarity of the query vector to the possible vectors in a database. Unstructured data like images, text, and audio can be represented as vectors within a high-dimensional space, aiming to efficiently locate and retrieve vectors that closely match the query vector. Metrics such as Euclidean distance or cosine similarity are commonly used to assess the similarity between vectors.

Euclidean distance between two vectors $\mathbf{P}$ and $\mathbf{Q}$ with two dimension is defined as follows.
\begin{equation}
\label{eq:euclidean}
  d(\mathbf{P},\mathbf{Q}) = \sqrt{(p_1 - q_1)^2 + (p_2 - q_2)^2}
\end{equation}

Cosine similarity between two vectors $\mathbf{P}$ and $\mathbf{Q}$ with n dimension is defined as follows.
\begin{equation}
\label{eq:cosine}
  \begin{split}
  \text{cosine similarity} &= cos(\theta) \\
  &= \frac{\mathbf{P} \cdot \mathbf{Q}}{\left\lVert \mathbf{P} \right\rVert \left\lVert \mathbf{Q} \right\rVert} \\
  &= \frac{\sum_{i=1}^{n} (p_i q_i)}{\sqrt{\sum{p_i^2}} \sqrt{\sum{q_i^2}}}
  \end{split}
\end{equation}

The constantly growing amount of available information resources has led to a high demand for efficient similarity searching methods. Vector search algorithms are designed to efficiently find the most similar vectors to a given query vector in a high-dimensional space. Some popular vector searching algorithms are K-Nearest Neighbors (KNN), Approximate Nearest Neighbor (ANN), and Product quantization (PQ).

Also known as the Brute force algorithm, KNN \cite{b12} finds the K closest vectors to the query vector by calculating the distance (often Euclidean distance) between the query and all other vectors in the dataset. It guarantees finding the exact nearest neighbors but can be computationally expensive for large datasets.

ANN \cite{b13} allows for a small error margin and returns points that are approximately nearest. This is not an exact nearest neighbors search, but an approximate closest points search. ANN loses some accuracy but gains a significant speedup compared to exact nearest neighbor search algorithms. Popular ANN approaches include k-d trees \cite{b14}, Locality Sensitive Hashing (LSH) \cite{b15}, and Hierarchical Navigable Small World Algorithm (HNSW) \cite{b16} .

K-d trees \cite{b14} are a type of binary search algorithm that efficiently organizes and searches large, multi-dimensional datasets. They work by recursively partitioning data along different dimensions at each level, using the median or mean to determine the splitting point. This creates a hierarchical structure that enables fast querying by traversing the tree and pruning branches based on the query point's coordinates, reducing the search space and improving efficiency.

Locality Sensitive Hashing (LSH) \cite{b15} is a technique that enables efficient similarity search by hashing similar points to the same bucket. Unlike traditional hashing, LSH maximizes collisions between similar points by applying multiple hash functions. This process involves hashing a dataset with multiple functions during preprocessing, and then hashing a query point with the same functions to retrieve similar candidate points. By doing so, LSH efficiently identifies points likely to be similar to a query point, making it effective for similarity search and nearest neighbor queries.

The Hierarchical Navigable Small World (HNSW) Algorithm \cite{b16} constructs a multi-layered graph where each node represents a data point. It navigates this graph hierarchically, starting from coarse to fine granularity, using edges to guide the search towards similar points and prune the search space. This approach balances efficiency and accuracy, making HNSW effective for finding nearest neighbors in large datasets.

While ANN algorithms can significantly speed up the search process, they have some limitations:
Approximate solution: ANN algorithms return an approximate solution instead of an exact solution.
Preprocessing: Creating an index or data structure requires additional computing, which can be even more expensive than the original search.
Scalability: The preprocessing step can be computationally expensive and may not scale well for large datasets.

Product quantization \cite{b32} is another searching algorithm that significantly compresses high-dimensional vectors, reducing memory size and accelerating nearest-neighbor searches. The process involves dividing a high-dimensional vector into equal-sized sub-vectors, each of which is then assigned an ID corresponding to the centroid it belongs to. This method can achieve impressive compression rates of over 95\%. However, it is a lossy compression technique, meaning that higher compression levels often come at the cost of decreased prediction accuracy. Additionally, product quantization requires clustering sub-vectors in a database and storing the centroid information of quantized buckets in memory. 

To address these limitations, our work focuses on finding an exact solution using KNN while reducing memory requirements by quantizing vectors. By representing vectors using a smaller number of bits, we can reduce the memory requirements and improve the efficiency of the search process.

\section{Methodology}
There have been extensively studied quantization for LLMs\cite{b22,b23,b24,b25,b26,b27}. We would like leverage some of these technologies.

\begin{figure}[ht]
\centerline{\includegraphics[width=0.8\linewidth]{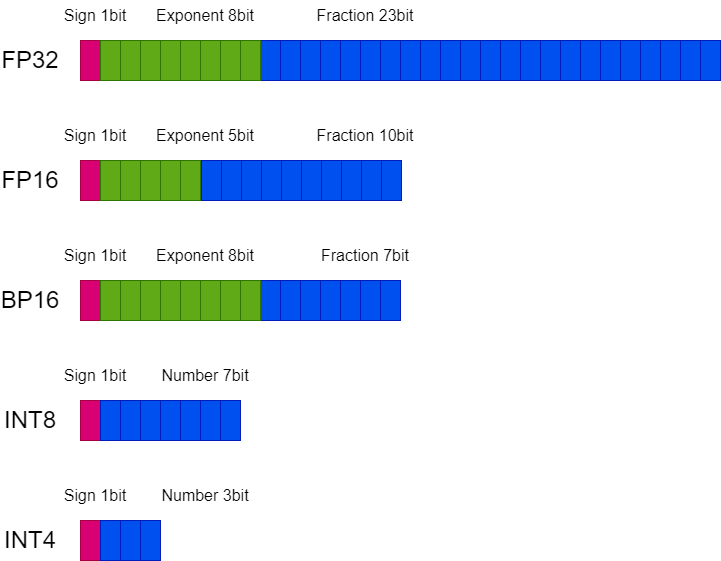}}
\caption{Data types}
\label{fig:datatype}
\end{figure}

\subsection{BFloat16}
Floating Point 32-bit (FP32) is a default data type for neural network computations and its weights and activation. It consists of 1 bit sign, 8 bits exponent, and 23 bits fraction. When precision calculation is not critical, Floating Point 16-bit (FP16) can be used since it reduces the memory and speed up communication. FP16 consists of 1 bit sign, 5 bits exponent, and 10 bits fraction. Recently Bfloat16 (Brain Floating Point 16-bit, BF16)\cite{b17} is becoming popular for machine learning tasks and its hardware implementation. BF16 is also 16 bits, but with a different format. It has 1 bit sign, 8 bits exponent, and 7 bits fraction. BF16 has a wider range but lower precision compared to FP16. BF16 provides the same range of FP32 and comes close to being a drop-in replacement for FP32 during training and deploying neural networks. In the case of FP16, special handling such as loss scaling might be required. BF16 is now supported in modern CPU, GPU, and NPU. In our work, default computing was conducted in BF16. Fig.\ref{fig:datatype} compares each datatype.     

\subsection{INT8}
INT8 uses 8 bits integers instead of floating point numbers and integer computation instead of floating point computing, reducing both memory and computing requirements. INT8 quantization has been applied to many neural networks to decrease the computational time and energy consumption of neural networks\cite{b18}. In neural network quantization, the weights and activation tensors are stored in 8-bit precision than the 32-bit precision which were trained. When transforming from 32 to 8 bits, the memory overhead of storing tensors decreases by a factor of 4 while the computational cost for matrix multiplication reduces by a factor of 16\cite{b19}. However, the use of 8-bit quantization in a network can result in quantization loss, which may subsequently cause a decrease in the model's overall accuracy. 

\subsection{Quantization process}
Suppose $x \in R^d$ and $x_q \in R^d$ are a full-precision and a quantized vector respectively.

Using the following symmetric linear quantization function\cite{b20}, a full-precision $x$ can be  mapped to a quantized vector $x_q$. 

\begin{equation}
\label{eq:quant}
  x_{q} = Clamp(Round[\frac{x}{s}])
\end{equation}

where $Clamp$ restricts the value of its argument to the quantized range $[-{2^{b-1}},{2^{b-1}-1}]$, $b$ is the number of bits for quantization. $\it{Round}$ is the round-to-nearest operator, and $S \in R$ is the scaling factor as defined as $S = \frac{max(|x|)}{2^{b}-1}$.

To invert from a quantized vector $x_q$ to a full-precision $x$, the scaling factor is simply multiplied.

\begin{equation}
\label{eq:dequant}
  x_{dq} = s\times x_{q}  
\end{equation}

The difference between $x$ and $x_{dq}$, $|x-x_{dq}|$ is called quantization loss.

\subsection{INT4 and Group-wise quantization}
While INT8 quantization is effective in reducing both the memory and computing cost while preserving accuracy in neural network and vector database, it remains unclear whether we can leverage INT4 to achieve further memory reduction. In this study, we explore the feasibility of employing INT4 quantization for vector space with group-wise quantization.
  
Directly quantizing high-dimensional vectors  as an entirety with the same quantization scale can significantly degrade the accuracy in INT4 quantization\cite{b19}. In the case of group-wise quantization\cite{b21}, a high-dimensional vector would be split into several groups with same group size. Each group can have its own quantization scale. 
The effect of group-wise quantization is further investigated in section \ref{Experiment}.

\section{Experiment}
\label{Experiment}


We designed two experimental setups to evaluate the accuracy of various quantization methods. The first setup uses cosine similarity scores to compare the precision of different data types, focusing on 4-bit quantization and group size effects. The second setup analyzes accuracy degradation in retrieval tasks, comparing our quantization methods to a 32-bit baseline. This comprehensive evaluation framework provides valuable insights into the trade-offs between quantization precision and accuracy.

\begin{figure}[ht]
\begin{center}
  \subfloat[][FP32]{\includegraphics[width=0.5\linewidth]{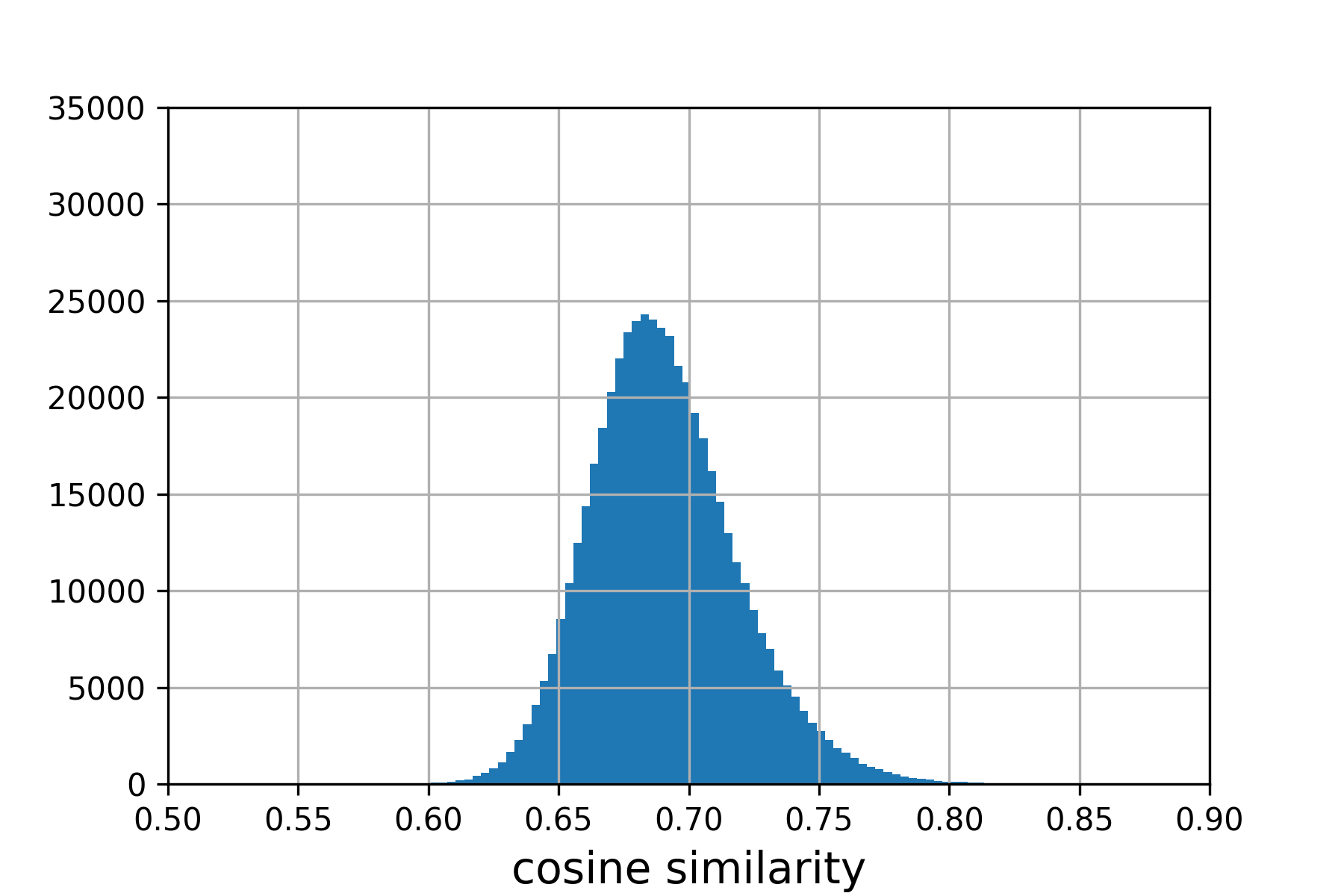}}
  \subfloat[][BF16]{\includegraphics[width=0.5\linewidth]{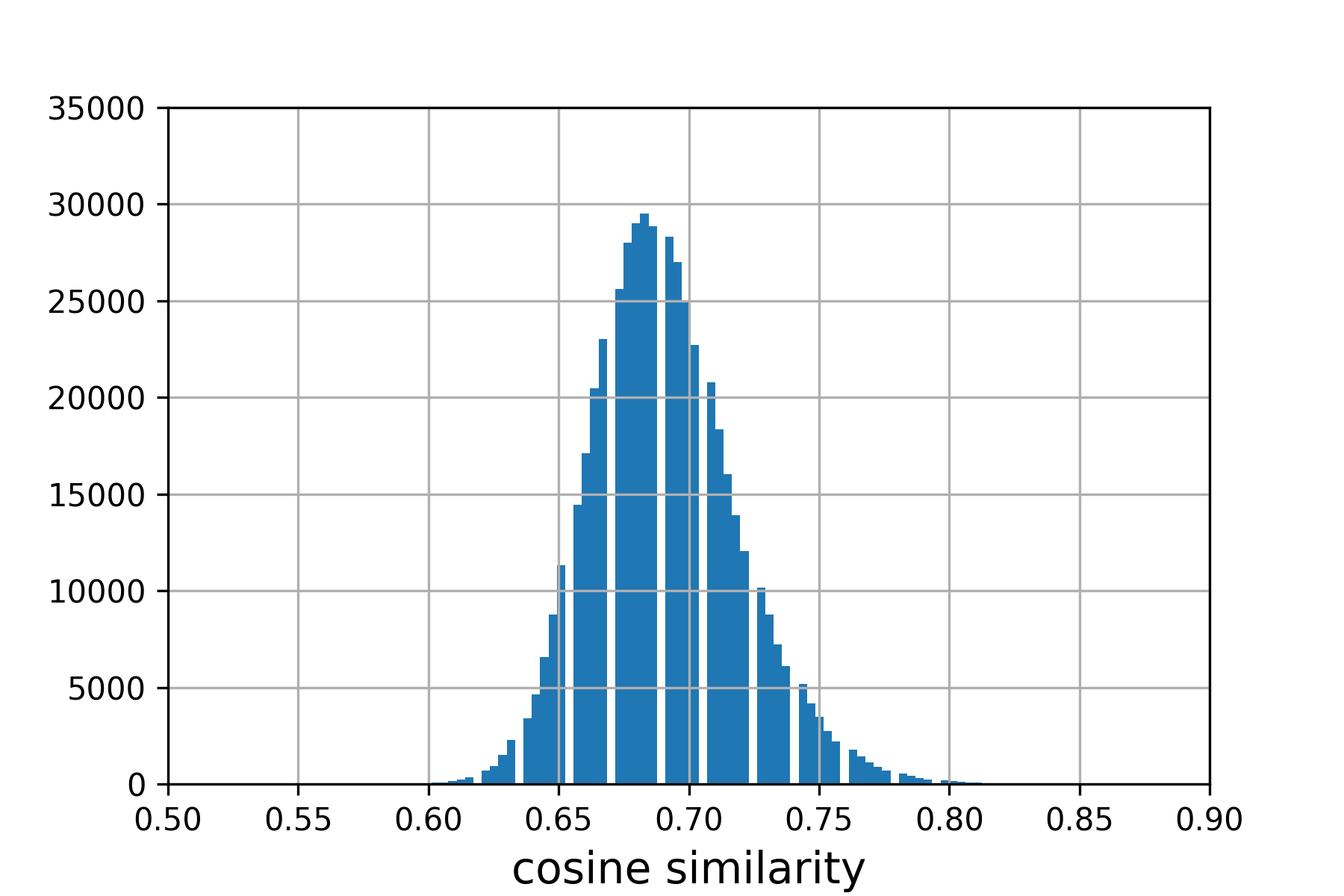}}
  \vfil
  \subfloat[][INT8]{\includegraphics[width=0.5\linewidth]{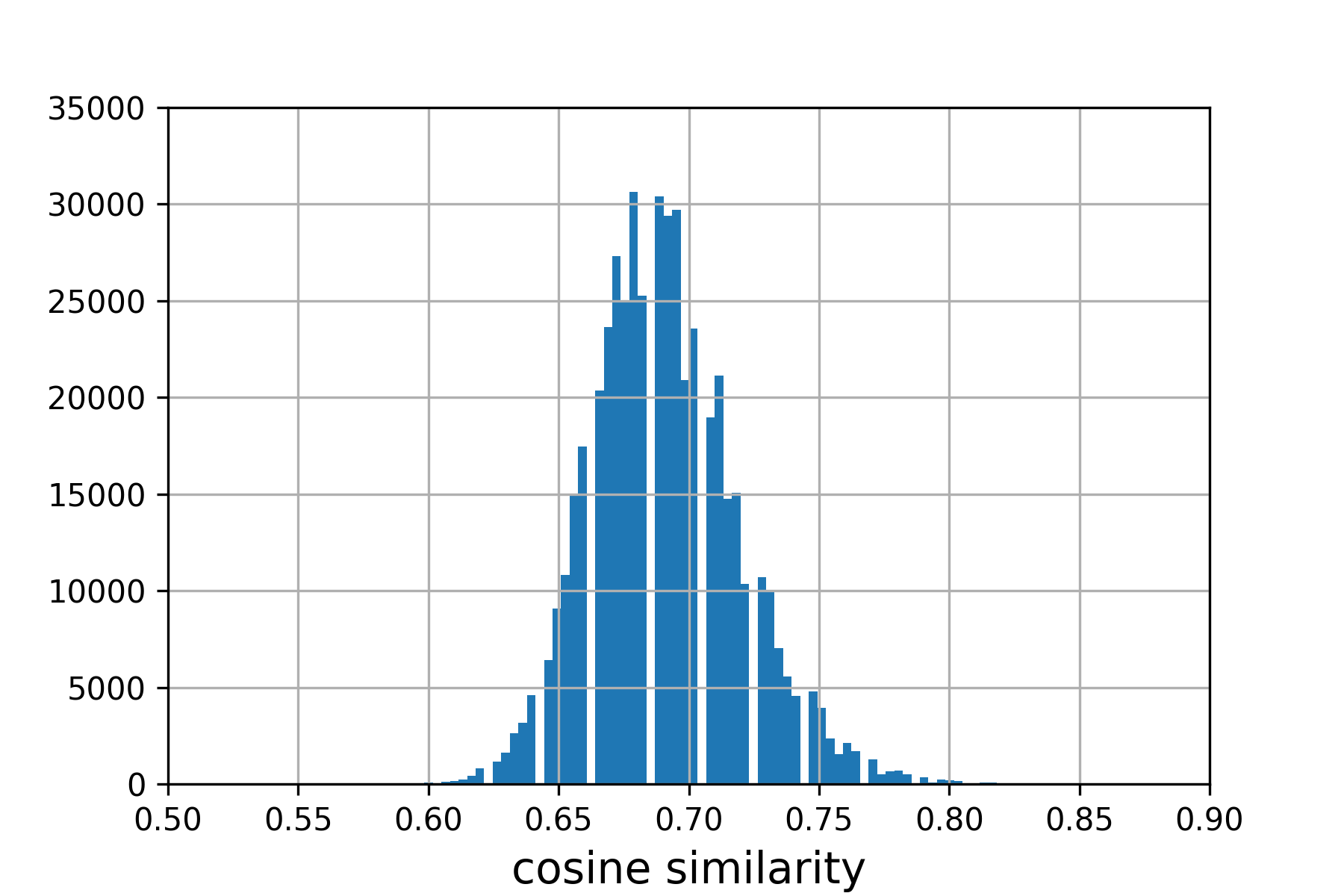}}
  \subfloat[][INT4]{\includegraphics[width=0.5\linewidth]{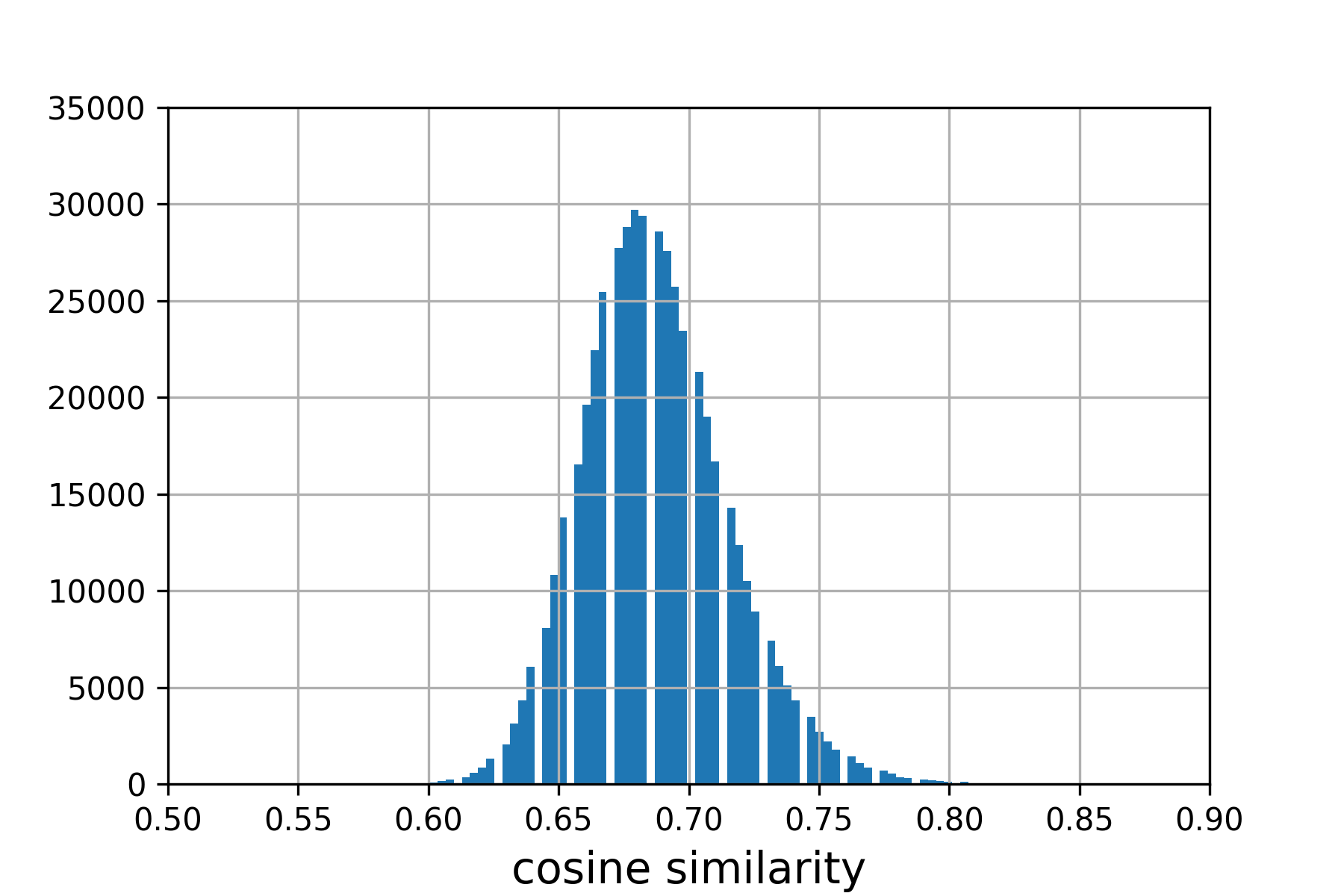}}

\end{center}
\caption{Distribution of cosine similarity for all pairs of 1000 vectors. (d) INT4 with group size 32} 
\label{fig:cosine}
\end{figure}

For the cosine similarity measurement, we randomly selected 1000 embedding vectors from dbpedia-openai-1M-1536-angular dataset\cite{b9}. Since the total number of available pairs is $\frac{N \times (N-1)}{2}$, we measured similarity score for $\frac{1000 \times 999}{2} = 499,500$ pairs of the embedding vectors. The distributions of the measured cosine similarity score of all the pairs of the 1000 vectors for different quantization method are shown in Fig \ref{fig:cosine}.

\begin{figure}[ht]
\centerline{\includegraphics[width=0.8\linewidth]{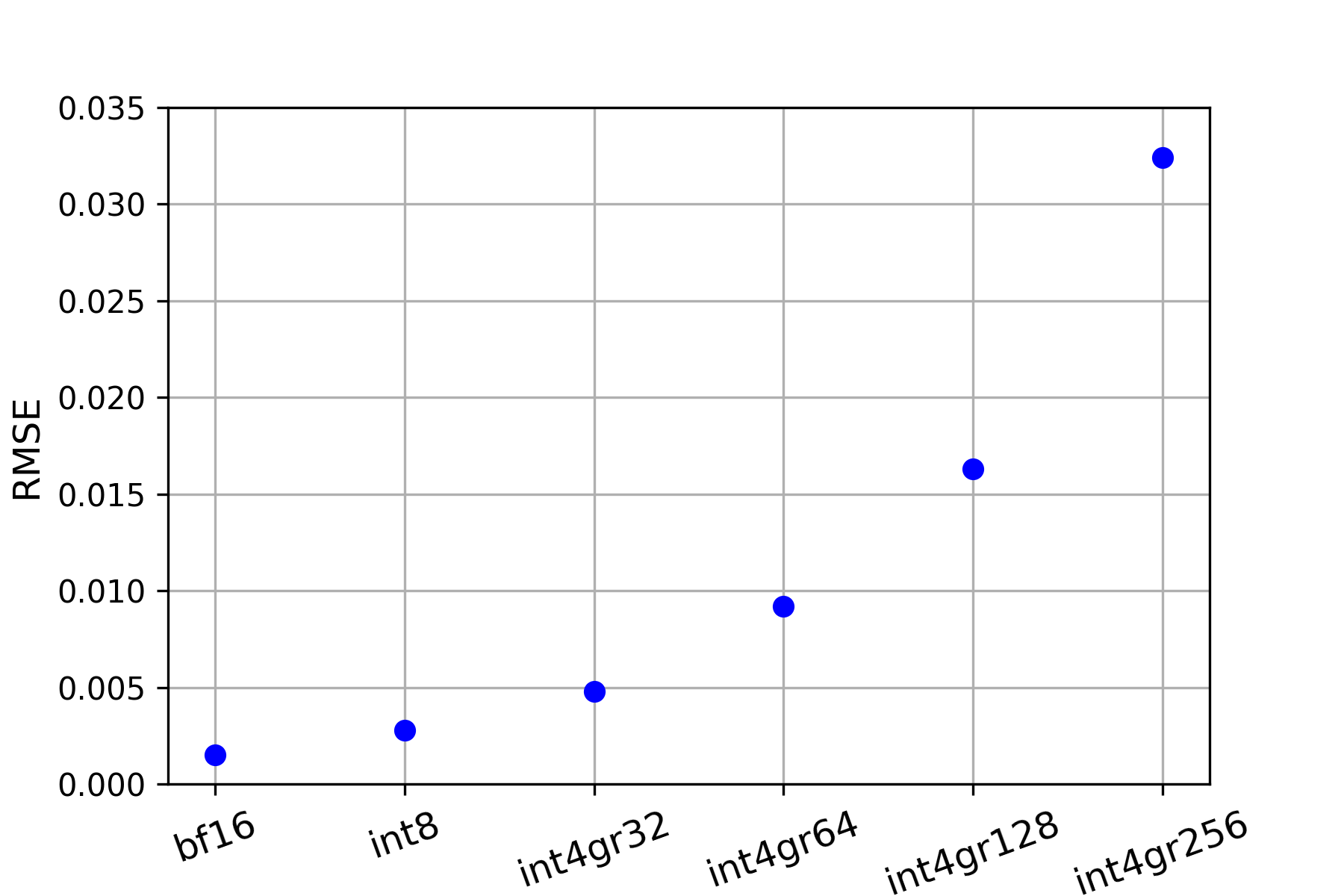}}
\caption{RMSE of the cosine similarity values of quantized pairs}
\label{fig:rmse_dt}
\end{figure}

Fig.\ref{fig:cosine} presents four histograms representing the distribution of cosine similarity values for different data representations: FP32, BF16, INT8, and INT4. All histograms exhibit a bell-shaped curve, indicating a normal distribution of cosine similarity values. 
The peak of the distribution shifts slightly to the left as the data representation moves from FP32 to INT4. This implies that the average cosine similarity decreases with lower precision data representations. The width of the distribution also increases as the data representation moves from FP32 to INT4. This indicates a larger spread of cosine similarity values for lower precision representations, suggesting potentially more variability in the data. The tails of the distributions become heavier as the data representation moves from FP32 to INT4. This means that there are more data points with extremely low or high cosine similarity values in the lower precision representations. The observed changes in the histograms suggest that quantizing the data from FP32 to lower precision formats (BF16, INT8, and INT4) introduces quantization loss and reduces the accuracy of the cosine similarity calculation. This can lead to a decrease in the average similarity and an increase in the variability of the results. This observation is consistent with the RMSE result in Fig.\ref{fig:rmse_dt}.

Root Mean Square Error (RMSE) between two vectors $\mathbf{P}$ and $\mathbf{Q}$ is defined as follows. 
\begin{equation}
\label{eq:rmse}
  \text{RMSE}(\mathbf{P}, \mathbf{Q}) = \sqrt{\frac{\sum_{i=1}^{n} (p_i - q_i)^2}{N}}
\end{equation}
To evaluate the accuracy of the cosine similarity values of quantized pairs, we calculated the RMSE between their cosine similarity values and those of the baseline 32-bit floating-point representations. 
The result is shown in Fig.\ref{fig:rmse_dt} and Table \ref{tab_rmse}. 


\begin{table}[ht]
\caption{RMSE of the cosine similarity values of quantized pairs}
\label{tab_rmse}
\begin{center} 
\scalebox{0.9}
{
\begin{tabular}{|l|c|c|c|}
\hline
\rule[-1ex]{0pt}{3.5ex}  Datatype & Group size & RMSE \\\hline\hline
\rule[-1ex]{0pt}{3.5ex}  Bf16 & {} & 0.0015 \\\hline
\rule[-1ex]{0pt}{3.5ex}  Int8 & {}& 0.0028 \\\hline
\rule[-1ex]{0pt}{3.5ex}  Int4 & 32 & 0.0048 \\\hline
\rule[-1ex]{0pt}{3.5ex}  Int4 & 64 & 0.0092  \\\hline
\rule[-1ex]{0pt}{3.5ex}  Int4 & 128 & 0.0163  \\\hline
\rule[-1ex]{0pt}{3.5ex}  Int4 & 256 & 0.0324  \\\hline
\end{tabular}\vspace{-20pt}
}
\end{center}
\end{table}

Fig.\ref{fig:rmse_dt} illustrates the relationship between various quantization methods and their respective error rate of the cosine similarity. Since the embedding vectors are normalized to a range of 0 to 1, the RMSE values can be interpreted as an error rate. The chart shows a clear trend: Error rates tend to rise as the datatype shifts from BF16 to INT4. 
The accuracy of quantization methods, as measured by cosine similarity, decreases as the datatype precision is reduced. For INT4, the accuracy decreases as the group size increases.

\begin{figure}[ht]
\includegraphics[width=\linewidth]{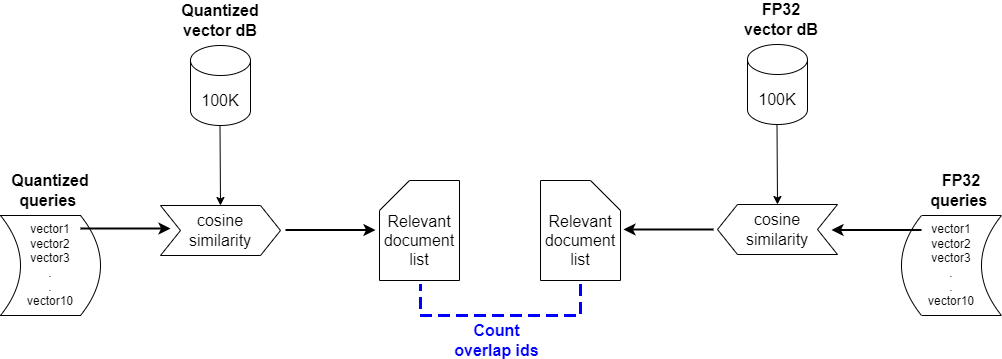}
\caption{Searching relevant documents and counting the overlaps}
\label{fig:relevant}
\end{figure}

To evaluate the accuracy of informational retrieval using our vector quantization approach, we conducted a comprehensive experiment. We split dbpedia-openai-1M-1536-angular dataset into 900K and 100K after permutation. Then, we carefully selected 10 embedding vectors from the 900K dataset. The magnitude of cosine similarity values for all possible pairs among the 10 selected vectors is less than 0.1, which means they are almost orthogonal to each other. 
Using the 10 selected vectors as queries, we searched relevant vectors from the 100K dataset. The overall process is illustrated in Fig.\ref{fig:relevant}

First, we computed the cosine similarity value between each query vector and every vector in the 100K dataset. This allowed us to identify the top 10 most relevant vectors for each query vector, based on their cosine similarity scores. 
This resulted in a total of 100 unique vectors that did not overlap with each other. We treated this list of 100 vectors as our baseline, representing the most relevant results without any quantization.

\begin{figure}[ht]
\centerline{\includegraphics[width=0.8\linewidth]{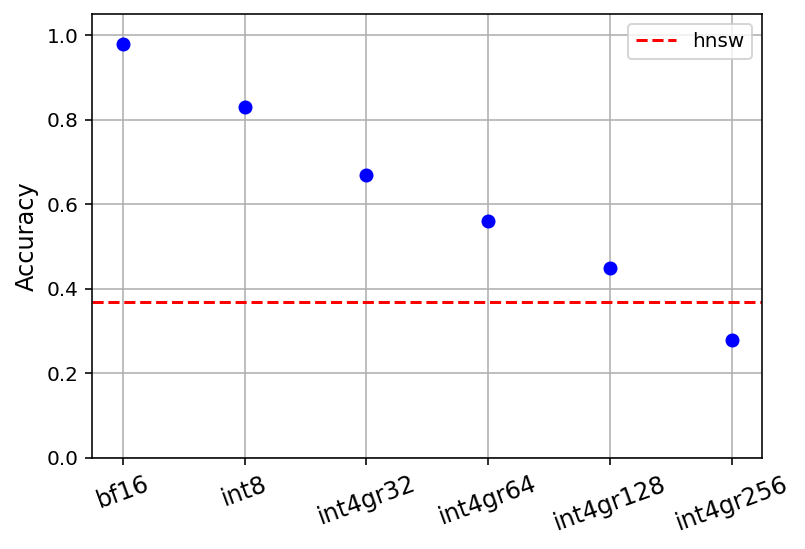}}
\caption{Accuracy of retrieval for quantization}
\label{fig:acc_dt}
\end{figure}

Next, we applied vector quantization to the entire vector space and then searched for the top 10 relevant vectors for each of the 10 query vectors, based on their cosine similarity scores. This time, however, we used the quantized vectors instead of the original floating-point vectors. We then compared the resulting list of top 10 relevant vectors for each query vector with the baseline list obtained earlier. To assess the accuracy of our quantization approach, we counted the number of vectors that overlapped between the two lists. In other words, we measured the proportion of vectors that were common to both the baseline list and the list obtained after quantization. This overlap metric served as a proxy for the accuracy of retrieval for quantization. The result is shown in Fig.\ref{fig:acc_dt} and Table \ref{tab_acc}. 

Fig.\ref{fig:acc_dt} illustrates the relationship between various quantization methods and their corresponding retrieval accuracy values. Notably, the dotted line represents the accuracy achieved by the HNSW algorithm (M=64 and ef=50) \cite{b16} which is a state-of-the-art algorithm used for an approximate search of nearest neighbours. The data reveals a downward trend, indicating that as the precision of the quantized vectors decreases, the accuracy values also tend to decrease. Although the accuracy drops significantly with INT4, using INT4 with a group size of 128 or less yields higher accuracy than the HNSW algorithm. 


\begin{table}[ht]
\caption{Accuracy of retrieval for quantization}
\label{tab_acc}
\begin{center} 
\scalebox{0.9}
{
\begin{tabular}{|l|c|c|c|}
\hline
\rule[-1ex]{0pt}{3.5ex}  Datatype & Group size & Accuracy \\\hline\hline
\rule[-1ex]{0pt}{3.5ex}  Bf16 & {} & 0.98 \\\hline
\rule[-1ex]{0pt}{3.5ex}  Int8 & {}& 0.83 \\\hline
\rule[-1ex]{0pt}{3.5ex}  Int4 & 32 & 0.67 \\\hline
\rule[-1ex]{0pt}{3.5ex}  Int4 & 64 & 0.56  \\\hline
\rule[-1ex]{0pt}{3.5ex}  Int4 & 128 & 0.45  \\\hline
\rule[-1ex]{0pt}{3.5ex}  Int4 & 256 & 0.28  \\\hline
\end{tabular}\vspace{-20pt}
}
\end{center}
\end{table}

\section{Comparative analysis}
\label{analysis}

We compared our quantization method with Product-quantization \cite{b32}. We conducted two experiments. First, we evaluated the accuracy of searching relevant documents after Product-quantization.  We quantized the 100K dbpedia-openai-1M-1536-angular dataset using Product-quantization algorithm. The experimented combinations of the number of sub-vectors, M and the number of quantized buckets, K as follows. 
$(48, 256), (32, 256), (16, 256), (8, 256), (32, 16), (16, 16),$
$(8, 16)$.
After computing the cosine similarity value between each query vector and every vector in the 100K dataset, we identified the top 10 most relevant vectors for each query vector as described in the 
 section \ref{Experiment}. We then compared the resulting list of top 10 relevant vectors after Product quantization with the baseline list from the original floating-point vectors. The overlap ratio of the two lists, which is the accuracy of retrieval was less than 0.1.  This suggests that Product Quantization loses its retrieval accuracy and is unable to identify the exact solution when searching for relevant vectors using cosine similarity.

To extend our comparison analysis with Product-quantization, we compared Pearson correlation coefficient between human-evaluated values and cosine similarity values from three Semantic textual similarity datasets such as sts-metb \cite{b28}, str-2022 \cite{b29}, and SICK \cite{b30}. These datasets have two pair of sentences and its semantic textual similarity or relatedness score, which was evaluated by humans. The size of each dataset is 8,628 for sts-metb, 5,500 for str-2022, and 9,840 for SICK. 

First of all, we split the each dataset into train (50\%) and test (50\%) dataset. Then, we embedded the pair of the sentences from each dataset using bge-large-en-v1.5 \cite{b31} embedding model. The dimension of the embedded vector is 1024. We quantized the embedded vectors of test dataset based on quantized train dataset using Product-quantization algorithm. Then, we evaluated the cosine similarity value between the pair of quantized vectors. Finally, we computed the correlation coefficient between the measured cosine similarity values and the semantic relatedness scores. 

\begin{table}[ht]
\caption{Correlation coefficients for Semantic Textual Similarity}
\label{tab_pc}
\begin{center} 
\scalebox{0.9}
{
\begin{tabular}{|l|c|c|c|c|}
\hline
\rule[-1ex]{0pt}{3.5ex}  Dataset & Datatype & Correlation coefficient & Ratio \\\hline\hline
\rule[-1ex]{0pt}{3.5ex}  sts-metb & Fp32 & 0.8830 & 1.0   \\\hline
\rule[-1ex]{0pt}{3.5ex}  sts-metb & Bf16 & 0.8827 & 0.9997 \\\hline
\rule[-1ex]{0pt}{3.5ex}  sts-metb & Int8 & 0.8461 & 0.9583 \\\hline
\rule[-1ex]{0pt}{3.5ex}  sts-metb & Int4 & 0.8474 & 0.9597 \\\hline
\rule[-1ex]{0pt}{3.5ex}  sts-metb & PQ[32, 256]\tablefootnote{Product-Quantization [M,K], 
   M: Number of sub-vectors, K: Number of quantized buckets} & 0.5970 & 0.6762 \\\hline
\rule[-1ex]{0pt}{3.5ex}  sts-metb & PQ[32, 16] & 0.5568 & 0.6305 \\\hline
\hline
\rule[-1ex]{0pt}{3.5ex}  str-2022 & Fp32 & 0.8212 & 1.0   \\\hline
\rule[-1ex]{0pt}{3.5ex}  str-2022 & Bf16 & 0.8212 & 1.0 \\\hline
\rule[-1ex]{0pt}{3.5ex}  str-2022 & Int8 & 0.7833 & 0.9538 \\\hline
\rule[-1ex]{0pt}{3.5ex}  str-2022 & Int4 & 0.7812 & 0.9513 \\\hline
\rule[-1ex]{0pt}{3.5ex}  str-2022 & PQ[32, 256] & 0.4041 & 0.4921 \\\hline
\rule[-1ex]{0pt}{3.5ex}  str-2022 & PQ[32, 16] & 0.4396 & 0.5353 \\\hline
\hline
\rule[-1ex]{0pt}{3.5ex}  SICK & Fp32 & 0.8637 & 1.0   \\\hline
\rule[-1ex]{0pt}{3.5ex}  SICK & Bf16 & 0.8633 & 0.9996 \\\hline
\rule[-1ex]{0pt}{3.5ex}  SICK & Int8 & 0.8266 & 0.9571 \\\hline
\rule[-1ex]{0pt}{3.5ex}  SICK & Int4 & 0.8285 & 0.9592 \\\hline
\rule[-1ex]{0pt}{3.5ex}  SICK & PQ[32, 256] & 0.6391 & 0.7400 \\\hline
\rule[-1ex]{0pt}{3.5ex}  SICK & PQ[32, 16] & 0.6358 & 0.7360 \\\hline
\end{tabular}\vspace{-20pt}
}
\end{center}
\end{table}

The table \ref{tab_pc} provides a detailed comparison of the correlation coefficient for different data types and quantization. Our quantized INT8 and INT4 demonstrate a slight degradation in coefficients, reaching up to 4\%. In contrast, Product-quantization exhibits a significant degradation, with coefficients declining by as much as 50\%.

\section{Limitation}
Converting embedding vectors from higher precision formats, such as 32-bit floating point, to lower precision formats like 8-bit or 4-bit integers, can lead to significant increases in searching speed. The key reason for this is that using fewer bits to represent data simplifies calculations and accelerates processing on hardware. However, to fully realize these speed gains, dedicated hardware instructions are necessary to support calculations with lower precision integer data. Currently, many popular frameworks, including PyTorch, only support the INT8 data type, but do not yet support INT4. To confirm the expected performance benefits, it would be essential to measure the actual impact on searching speed, which was not done in this work.

\section{Conclusion}
Recently, RAG has emerged as a promising solution to the limitations of LLMs, particularly the issues of outdated information and hallucinations. RAG uses large-scale similarity search to retrieve relevant information from extensive databases of images, audio, video, and text. The key process of RAG pipeline is storing information as high-dimensional embedding vectors, 
which require substantial memory. This poses a challenge for deploying RAG models on devices with limited memory, such as mobile phones.

To address this challenge, we utilized quantization techniques to lower the precision of these high-dimensional vectors. This quantization can reduce the memory demands of the database by a factor of four or eight. Our findings indicate that 8-bit quantization maintains retrieval accuracy with only slight degradation. Group-wise quantization can alleviate some of the accuracy loss encountered with 4-bit quantization. Furthermore, the improved memory efficiency of 4-bit quantization allows for the use of larger vector databases, which could potentially produce more accurate and relevant search results.

\vspace{12pt}
\color{red}

\end{document}